\pdfoutput=1
\documentclass[journal]{IEEEtran}

\usepackage{lineno,hyperref}
\usepackage{float}
\usepackage{graphicx}
\usepackage{adjustbox}
\usepackage{color,xcolor}
\usepackage{booktabs}
\usepackage{multirow}
\usepackage{stfloats}
\usepackage{lipsum,cuted}
\usepackage{float}
\usepackage{caption}
\usepackage{tabularx}
\usepackage{multirow}
\usepackage{array}

\usepackage{subfigure}
\usepackage{ulem}
\usepackage{svg}
\usepackage{amsmath}
\usepackage{amsthm,amsmath,amssymb}
\usepackage{mathrsfs}

\ifCLASSINFOpdf
\else
\fi

\begin{document}
%


\title{Adaptive Graph Convolutional Network Framework for Multidimensional Time Series Prediction}

\author{Ning~Wang,~
       
\thanks{\emph{(Corresponding author: Ning~Wang)}}
\thanks{

}

}

%
%

\markboth{Journal of Selected Topics in Applied Earth Observations and Remote Sensing,~Vol.~XX, No.~X, XX~2021}%
{Shell \MakeLowercase{\textit{et al.}}: Bare Demo of IEEEtran.cls for IEEE Journals}
%



\maketitle

\begin{abstract}
In the real world, long sequence time-series forecasting (LSTF) is needed in many cases, such as power consumption prediction and air quality prediction.Multi-dimensional long time series model has more strict requirements on the model, which not only needs to effectively capture the accurate long-term dependence between input and output, but also needs to capture the relationship between data of different dimensions.Recent research shows that the Informer model based on Transformer has achieved excellent performance in long time series prediction.However, this model still has some deficiencies in multidimensional prediction,it cannot capture the relationship between different dimensions well.
We improved Informer to address its shortcomings in multidimensional forecasting.
First,we introduce an adaptive graph neural network to capture hidden dimension dependencies in mostly time series prediction. Secondly,we integrate adaptive graph convolutional networks into various spatio-temporal series prediction models to solve the defect that they cannot capture the relationship between different dimensions. Thirdly,After experimental testing with multiple data sets, the accuracy of our framework improved by about 10\% after being introduced into the model.

\end{abstract}


%
\IEEEpeerreviewmaketitle

\section{Introduction}
%
%
%
%

Time-series prediction plays an important role in many fields, such as sensor network monitoring\cite{papadimitriou2006optimal},energy and smart grid management,economics and finance\cite{zhu2002statstream},and disease propagation analysis\cite{matsubara2014funnel}. In each of the previous scenarios, we can make long-term predictions using a large number of past time series,namely long sequence time-series forecastin(LSTF). Multidimensional time series prediction is an important part of it. The so-called multidimensional time series prediction refers to the generation of multiple kinds of data at the same time, using this datas at the historical moment to predict the data at the future moment. This paper mainly discusses the different types of time series data detected in the same place at the same time. However, existing models cannot learn the hidden relationship between different dimensions well. In the previous period,vector autoregression(VAR) is arguably the most widely used models in multivariate time series\cite{box2015time}\cite{hamilton2020time}\cite{lutkepohl2005new} due to its simplicity. In recent years, various VAR models have made significant progress,including the elliptical VAR\cite{qiu2015robust} model for heavy-tail time series and structured VAR\cite{melnyk2016estimating} model in order to better explain the dependence between high dimensional variables. However,the model capacity of VAR grows linearly over the temporal window size and quadratically over the number of variables. This means that the model is easy to overfit when it deals with long time series. In order to alleviate this problem,\cite{yu2016temporal}proposed to reduce the original high-dimensional signal to low-dimensional implicit representation, and then use VAR to perform a variety of regularization prediction.

The time series prediction problem can also be regarded as the standard regression problem with time-varying parameters. Therefore, there is a lot of work to apply various regression models with different loss functions and regularization terms to time series prediction tasks. For example,linear support vector regression(SVR) learns a hyperplane based on regression loss and controls the threshold of prediction error with the hyperparameter E.Lastly,\cite{li2014forecasting} used LASSO models to encourage sparsity in the model parameters so that interesting patterns among different input signals can be manifest. These linear methods are actually more efficient in multivariate time series prediction due to high-quality off-the-shelf solvers in the machine learning community. However,like VAR,these models cannot capture the complex nonlinear relationship between multiple dimensions, resulting in low performance. And the power of GP comes at the cost of high computational complexity. Due to the inverse of kernel matrix, the prediction of multivariate time series using this model has cubic complexity. 

Gaussian Processes(GP) is a nonparametric method for modeling distributions over a continuous domain of functions. GP can be applied to the multivariate time series prediction task suggested in \cite{roberts2013gaussian}. For example,\cite{frigola2013bayesian}proposes a full Bayesian method with GP priors for nonlinear state space models, which can capture complex dynamic phenomena.

In the LSTNet model\cite{lai2018modeling}, the convolutional network layer used cannot solve this problem well.As for the model LSTMa originally used for text translation\cite{bahdanau2014neural}, its Recurrent Neural Network module cannot deal with the problem of multi-dimensional dependence well when it is transferred to time series prediction. There is another serious problem with the model based on LSTM.  \cite{zhou2021informer} points out that the accuracy of LSTM prediction will decrease greatly when the prediction length reaches a certain length. Therefore, the model does not perform well in the prediction of long time series.

At present, there are two main challenges in multi-dimensional long time series prediction. The first is that the model should be able to deal with the increasingly long prediction series. Second, as there are dependencies between different dimensions in multidimensional sequence data, the model needs to capture this dependencies. Recently emerged Transformer models perform well in the task of capturing long-range dependency than recurrent neural networks models(RNN). The self-attention mechanism can reduce the transmission distance of signals from the maximum length to the minimum $\mathcal{O}(1)$ and avoid the recurrent structure, thus Transformer has greater potential than RNN in dealing with LSTF problem. However, due to its \emph{L}-quadratic computation and memory consumption on \emph{L}-length inputs/outputs of Transformer model, it has a large cost for long time series prediction. In order to solve these problems,Reformer\cite{kitaev2020reformer} achieves $\mathcal{O}(L \log L)$ with locally-sensitive hashing self-attention,but it only works on extremely long sequences. Linformer claims a linear complexity $\mathcal{O}(L)$,but the project matrix can not be fixed for real-world long sequence input, which may have the risk of degradation to $\mathcal{O}\left(L^{2}\right)$. Transformer-XL\cite{dai2019transformer} and Compressive Transformer\cite{rae2019compressive} use auxiliary hidden states to capture long-range dependency. These efforts are mainly focused on the quadratic computation of self-attention,thus neglecting the memory bottleneck in stacking layers for long inputs which limits the model scalability in receiving long sequence inputs. 

 \cite{zhou2021informer} proposed ProbSparse self-attention mechanism and self-attention distilling on this basis, which greatly reduced the time and space complexity of the model. Since then,one pressing demand
is to extend the forecast time into the far future, which is quite meaningful for the long-term planning and early warning. 

However, this model only solves the first problem and still cannot model the dependency between different dimensions. 

Recently, graph convolutional neural(GCN) networks have shown superior performance in capturing spatial dependencies compared to convolutional neural networks(CNN). In GCN network, when updating node information, the updated information can be obtained by weighted summation of the node information around the node and the node itself.As is shown in\ref{fig:one}. In the spatiotemporal graph neural network model, GCN generally models sensors scattered in different Spaces, and each sensor corresponds to a graph node. So GCN can aggregate these spatial information by using known adjacency matrix. It occurred to us that since GCN was able to model different sensors, could sensors be replaced with each dimension of the multidimensional data? So that raises the question, what about the adjacency matrix? Because in the sensor scenario, the adjacency matrix is constructed according to the distance, if the two sensors are close enough, we can set the corresponding position of the adjacency matrix to 1. Inspired by Wu\cite{wu2019graph} et al. 's work, we know that the adjacency matrix can be learned by the model itself. So we thought, since we can't calculate the dependencies between different dimensions in advance, then we can make the model learn this dependency on its own. The nodes used to correspond to sensors at different locations, but now we replace the sensors with one of the multiple dimensions. 

To this end,Our work addresses the above multidimensional prediction problem.
We find that almost all the multidimensional time prediction models do not take into account the dependence between different dimensions,and we solved this problem by introducing self-adaption graph convolution network,and conduct extensive experiments. The contributions of this paper are summarized as follows:

\begin{figure}
	\centering
	\includegraphics[scale=0.3]{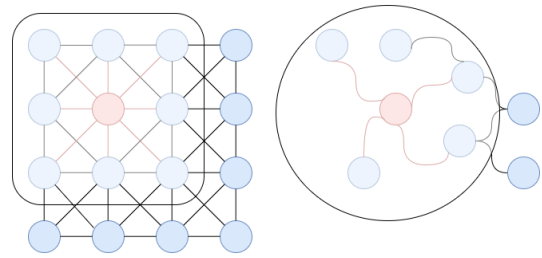}
	\caption{Left:This is the process of information aggregation in convolutional network.CNN transmits the information of the surrounding nodes to the red node.Right:This is how graph convolutional networks aggregate information.The biggest difference between two methods is that the network structure on the right is more free, and even different nodes can be in the same plane, while the structure on the left must be arranged in the norm.} 
	\label{fig:one}
\end{figure}

\begin{itemize}
	\item{We introduce adaptive GCN into the multi-dimensional time series prediction model so that the model can deal with the dependence between different dimensions.}
	\item{We took Adaptive GCN as a framework alone and fused it with the existing multidimensional time series model.}
	\item{We tested the fused models on several recognized data sets and got good result (about 10\%)s.}
\end{itemize}

\begin{figure}
	\centering
	\includegraphics[scale=0.3]{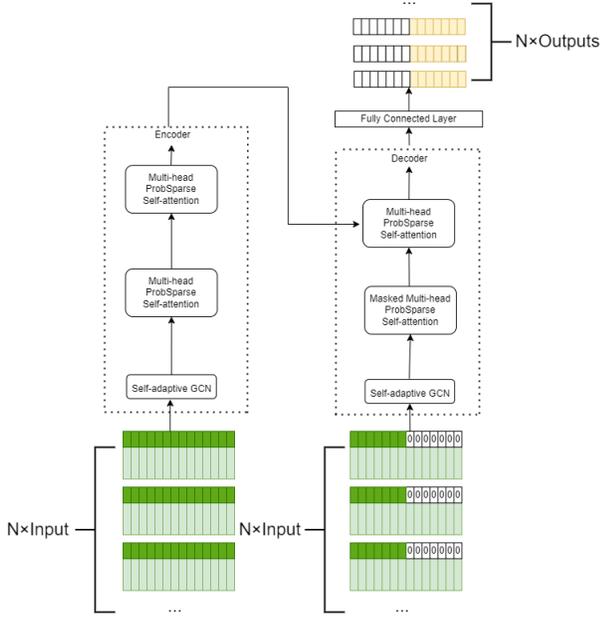}
	\caption{GCN-Informer model overview. Left:The encoder receives time series data of N dimensions. Before entering the decoder, it first learns the dependence of different dimensions through the stacked GCN layer (model details are given later), and then enters the encoder layer of Informer for information extraction of time features.Right:The same GCN operation is required to aggregate information from different dimensions before entering the decoder.The decoder receives the result of the previous layer and pads the target elements into zero,calculates the weighted attention composition of the feature map.Finally, output results(orange series) through the full connection layer.} 
	\label{fig:two}
\end{figure}

\ref{fig:two}

\section{Method}
In this section, we first give a mathematical definition of the problem we are trying to solve. Next, we will introduce two components of the framework, graph convolution layer and Attention layer,which GCN needs to be mainly described. Finally, summarize the overall framework model.

\subsection{Problem Definition}
The multidimensional time series data prediction problem is defined on the graph $G=(V,E,\mathbf{A})$,where $V$ is a set of nodes and $E$ is a set of edges. The adjacency matrix is $\mathbf{A}\in\mathbb{R}^{N\times N}$. If $v_i,v_j\in{E}$,then $A_{ij}$ is one otherwise it is zero. If you want $A_{ij}$ more precise representation of dependencies between nodes, it can be in $[0,1]$.The closer this value is to 1, the greater the dependence between i and j. In each time step $t$ , $G$ has a dynamic feature matrix $\mathbf{X}^{(t)}\in\mathbb{R}^{N\times M}$. Given H step graph signals of information in $G$,our problem is to learn a function f to predict the next T step graph signals as
\begin{equation}
	\left[\mathbf{X}^{(t-S): t}, G\right] \stackrel{f}{\rightarrow} \mathbf{X}^{(t+1):(t+T)},
\end{equation} 
where $\mathbf{X}^{(t-H):t}\in\mathbb{R}^{N\times M\times H}$ and $\mathbf{X}^{(t+1):(t+T)}\in\mathbb{R}^{N\times M\times T}$.

\subsection{Graph Convolution Layer}
At present, graph convolutional network is a very important model, which can extract node features with given node structure information.Kipf et al.\cite{kipf2016semi} proposed a first approximation of Chebyshev spectral filter\cite{defferrard2016convolutional}. Graph convolutional network has two ways based on spectrum or space. From the point of view of space, it is to smooth the signal of node by gathering and transforming the neighborhood information of node.The advantage of this method is that multidimensional data can be modeled and the prediction effect can be improved through aggregation and transfer mechanisms. $\tilde{\mathbf{A}}\in \mathbf{R}^{N\times N}$ represents the self-loops adjacency matrix. By adding the identity matrix, the graph convolution network can retain its own characteristic information during aggregation. In other words, $\mathbf{A}$ becomes $\tilde{\mathbf{A}}$ by adding an identity matrix. $\mathbf{X} \in \mathbf{R}^{N \times D}$represents the input signal. $\mathbf{X} \in \mathbf{R}^{N \times M}$represents the input output,$\mathbf{Z} \in \mathbf{R}^{N \times M}$,and $\mathbf{W} \in \mathbf{R}^{D \times M}$represents the model parameter matrix,in \cite{kipf2016semi} the graph convolution layer is defined as
\begin{equation} 
	\mathbf{Z}=\tilde{\mathbf{A}} \mathbf{X} \mathbf{W}.
\end{equation} 

\cite{li2017diffusion}proposed a diffusion convolution layer and integrated it into the recurrent neural network to make the model more effective in predicting spatio-temporal data. We summarise that the diffusion convolution layer into equation 3,which results in,
\begin{equation} 
	\mathbf{Z}=\sum_{k=0}^{K} \mathbf{P}^{k} \mathbf{X} \mathbf{W}_{\mathbf{k}}.
\end{equation}

\subsection{Self-adaptive Adjacency Matrix}
Wu et al.\cite{wu2019graph} proposed Graph Wavenet model to solve the prediction problem of spatio-temporal data, in which they introduced adaptive adjacency matrix into graph convolution network, so that the model did not need adjacency matrix based on spatial relations. This model can also incorporate adjacency matrix based on spatial distance. What they do is they operate on the adaptive adjacency matrix and the distance-based adjacency matrix separately and stack the results together. We know that in multidimensional prediction problems, there are dependencies between data of different dimensions, just like there are dependencies between different locations of spatio-temporal data. Therefore, since the adaptive adjacency matrix can be used to extract the dependencies of different spatial positions, it can also extract the dependencies between different dimensions. In air quality forecasts, for example, visibility should be strongly correlated with wind speed, since fog is only present when wind speeds are low. Then in the adjacency matrix, the intersection value of these two dimensions should be larger than that of other unrelated dimensions. In the spatio-temporal data, we can define the dependence of nodes in the adjacency matrix according to the distance between nodes, but in the multidimensional prediction problem, it is difficult to quantify the relationship between different dimensions. So we use this adaptive adjacency matrix and let the model learn this dependency by itself. The adaptive adjacency matrix is constructed by multiplying two learnable parameters. It is defined as
\begin{equation} 
	\tilde{\mathbf{A}}_{a d p}=\operatorname{SoftMax}\left(\operatorname{ReLU}\left(\mathbf{E}_{1} \mathbf{E}_{2}^{T}\right)\right) .
\end{equation} 
After combining the formula of adaptive adjacency matrix and graph convolution network, it can be defined as
\begin{equation} 
	\mathbf{Z}=\sum_{k=0}^{K} \tilde{\mathbf{A}}_{a p t}^{k} \mathbf{X} \mathbf{W}_{k}.
\end{equation} 

\subsection{Temporal Convolution Layer}
The adaptive GCN framework can be integrated into almost all multi-dimensional time series prediction models. Here, one of the models is taken as an example.
We adopt the Informer model\cite{zhou2021informer} based on Transformer structure to capture the time trends in each dimension. The Informer model improves the Transformer model by reducing the time and space complexity through ProbSparse self-attention mechanism, and enables the model to receive long sequences of inputs through self-attention distilling operation.And the model is a one-time output sequence to avoid cumulative errors.Informer processes sequence data in a non-recursive way compared to the RNn-based approach, which alleviates the problem of gradient explosions. Informer prevents the model from knowing the predicted result by setting the last few points of the decoder's input sequence to zero.

\section{Experiments}
\textbf{Datasets}

We conducted experiments on all four data sets, two of which were public data sets and the other two were real data sets.

\textbf{Weather}\footnote{Weather dataset was acquired at https://www.ncei.noaa.gov/data/local-climatological-data/.}:The weather data set includes local U.S. climate data from nearly 1,600 sites, collected at hourly intervals from 2010 to 2013.Each piece of data has 12 dimensions, of which 11 are the characteristics of the input and the remaining one is the target value with the prediction.

\textbf{ECL}(Electricity Consuming Load)\footnote{ECL dataset was acquired at https://archive.ics.uci.edu/ml/datasets/ElectricityLoadDiagrams20112014.}This data set collected the power consumption of 321 customers\cite{li2019enhancing}.We set characteristic MT320 as the target value.

\textbf{ETT}(Electricity Transformer Temperature)\footnote{ETT is posted on Github.https://github.com/zhouhaoyi/ETDataset}:The data set was collected over two years from two different counties in China. Data sets with an interval of 1 hour and 15 minutes are created respectively. Each data contains six power load characteristics and a target value "oil temperature".

\textbf{Experimental Details} 

\textbf{Baselines:}Since we are mainly improving the processing ability of the multidimensional data of the Informer model, We generally compare the original model with the model incorporating adaptive GCN. We choose the most advanced time series prediction models, including Autoformer\cite{xu2021autoformer}, Informer\cite{zhou2021informer}, Transformer\cite{vaswani2017attention}, Reformer\cite{kitaev2020reformer} and SCINet\cite{liu2021time} models. 

\textbf{Hyper-parameter tuning:}Again, take informer,informer contains 3-layer stack and a 1-layer stack in the encoder,and a 2-layer int the decoder. For the optimizer, we chose Adam, whose starting value of learning rate is $1 e^{-4}$,and the value of each epoch would be halved.Train-epochs for the model was set to 6 and patience was set to 3.Patience 3 means that if the loss is not minimized for three consecutive times, the training will be stopped.The training batch of the experiment is 32. In actual experiments, we generally do not change the parameters of the original model, and only compare the results with those after adding adaptive GCN, so as to avoid interference from other factors.

\textbf{Setup:}The input of each dataset is zero-mean normalized. In order to verify the effectiveness of GCN-Informer in long time series prediction, we also keep consistent with the prediction length of Informer's experiment. And the lengths are 24,48,168,336 respectively.There are other models that increase the predicted length even further, such as the minimum predicted length of 96 in the Autoformer article, Our predicted length is also consistent with it. The predicted length of 720 cannot be tested due to performance issues with the graphics card. \textbf{Metrics:} We use MSE and MAE criteria to test the effectiveness of the model.Its calculation formula is $\mathrm{MSE}=\frac{1}{n} \sum_{i=1}^{n}(\mathbf{y}-\hat{\mathbf{y}})^{2}$ and $\operatorname{MAE}=\frac{1}{n} \sum_{i=1}^{n}|\mathbf{y}-\hat{\mathbf{y}}|$. \textbf{Platform:} All experiments were done on a GTX1660super 6GB GPU.

\begin{table*}\centering
	\begin{tabular}{@{}|l|ll|ll|ll|ll|@{}}
		\toprule
		dataset &
		\multicolumn{2}{l|}{WTH} &
		\multicolumn{2}{l|}{ECL} &
		\multicolumn{2}{l|}{ETTh1} &
		\multicolumn{2}{l|}{ETTm1} \\ \midrule
		Model &
		\multicolumn{1}{l|}{Informer} &
		AdpInformer &
		\multicolumn{1}{l|}{Informer} &
		AdpInformer &
		\multicolumn{1}{l|}{Informer} &
		AdpAutoformer &
		\multicolumn{1}{l|}{Informer} &
		AdpInformer \\ \midrule
		Metric &
		\multicolumn{1}{l|}{MSE MAE} &
		MSE MAE &
		\multicolumn{1}{l|}{MSE MAE} &
		MSE MAE &
		\multicolumn{1}{l|}{MSE MAE} &
		MSE MAE &
		\multicolumn{1}{l|}{MSE MAE} &
		MSE MAE \\ \midrule
		24 &
		\multicolumn{1}{l|}{0.326 0.380} &
		\textbf{0.311 0.363} &
		\multicolumn{1}{l|}{0.239 0.347} &
		\textbf{0.237 0.338} &
		\multicolumn{1}{l|}{0.604 0.573} &
		\textbf{0.599 0.570} &
		\multicolumn{1}{l|}{0.344 0.393} &
		\textbf{0.358 0.406} \\ \midrule
		68 &
		\multicolumn{1}{l|}{0.414 0.441} &
		\textbf{0.389 0.435} &
		\multicolumn{1}{l|}{0.265 0.363} &
		\textbf{0.253 0.350} &
		\multicolumn{1}{l|}{0.692 0.630} &
		\textbf{0.650 0.597} &
		\multicolumn{1}{l|}{0.535 0.505} &
		\textbf{0.513 0.500} \\ \midrule
		168 &
		\multicolumn{1}{l|}{0.581 0.544} &
		\textbf{0.548 0.532} &
		\multicolumn{1}{l|}{\textbf{0.285 0.377}} &
		0.314 0.390 &
		\multicolumn{1}{l|}{\textbf{0.930} 0.751} &
		0.933 \textbf{0.750} &
		\multicolumn{1}{l|}{0.644 0.583} &
		\textbf{0.637 0.570} \\ \midrule
		336 &
		\multicolumn{1}{l|}{0.630 0.590} &
		\textbf{0.572 0.561} &
		\multicolumn{1}{l|}{0.309 0.397} &
		\textbf{0.305 0.381} &
		\multicolumn{1}{l|}{1.120 0.847} &
		\textbf{1.108 0.845} &
		\multicolumn{1}{l|}{0.931 0.734} &
		\textbf{0.884 0.706} \\ \bottomrule
	\end{tabular}
\end{table*}\centering

\begin{table*}\centering
	\begin{tabular}{@{}|l|ll|ll|ll|ll|@{}}
		\toprule
		dataset &
		\multicolumn{2}{l|}{WTH} &
		\multicolumn{2}{l|}{ECL} &
		\multicolumn{2}{l|}{ETTh1} &
		\multicolumn{2}{l|}{ETTm1} \\ \midrule
		Model &
		\multicolumn{1}{l|}{Autoformer} &
		AdpAutoformer &
		\multicolumn{1}{l|}{Autoformer} &
		AdpAutoformer &
		\multicolumn{1}{l|}{Autoformer} &
		AdpAutoformer &
		\multicolumn{1}{l|}{Autoformer} &
		AdpAutoformer \\ \midrule
		Metric &
		\multicolumn{1}{l|}{MSE MAE} &
		MSE MAE &
		\multicolumn{1}{l|}{MSE MAE} &
		MSE MAE &
		\multicolumn{1}{l|}{MSE MAE} &
		MSE MAE &
		\multicolumn{1}{l|}{MSE MAE} &
		MSE MAE \\ \midrule
		96 &
		\multicolumn{1}{l|}{0.590 0.552} &
		\textbf{0.492 0.503} &
		\multicolumn{1}{l|}{\textbf{0.201} 0.316} &
		0.217 \textbf{0.307} &
		\multicolumn{1}{l|}{0.448 \textbf{0.452}} &
		\textbf{0.441} 0.455 &
		\multicolumn{1}{l|}{0.542 0.492} &
		\textbf{0.422 0.438} \\ \midrule
		192 &
		\multicolumn{1}{l|}{0.612 0.567} &
		\textbf{0.552 0.543} &
		\multicolumn{1}{l|}{0.217 0.329} &
		\textbf{0.209 0.321} &
		\multicolumn{1}{l|}{0.501 0.485} &
		\textbf{0.462 0.465} &
		\multicolumn{1}{l|}{0.603 0.517} &
		\textbf{0.490 0.483} \\ \midrule
		336 &
		\multicolumn{1}{l|}{0.636 0.580} &
		\textbf{0.578 0.558} &
		\multicolumn{1}{l|}{0.319 0.413} &
		\textbf{0.215 0.328} &
		\multicolumn{1}{l|}{0.515 0.493} &
		\textbf{0.495 0.484} &
		\multicolumn{1}{l|}{0.609 0.524} &
		\textbf{0.576 0.523} \\ \midrule
		720 &
		\multicolumn{1}{l|}{0.664 0.598} &
		\textbf{0.615 0.583} &
		\multicolumn{1}{l|}{out of memory} &
		out of memory &
		\multicolumn{1}{l|}{0.552 0.536} &
		\textbf{0.613 0.555} &
		\multicolumn{1}{l|}{out of memory} &
		out of memory \\ \bottomrule
	\end{tabular}
\end{table*}\centering

\textbf{Results and Analysis}
Table \uppercase\expandafter{\romannumeral1} show the comparison of the results of multi-dimensional long time series prediction based on the original Informer model and the improved GCN-Informer model. We also restored the predicted lengths in the informer paper's experiment as much as possible, which were 24,48,168,336 respectively. Due to insufficient GPU memory, there is no test to predict a length of 720. Both MSE and MAE were calculated by means of 6 repetitions. The time was also the average of six repeated experiments. The best results are in bold. Taking the WTH data set as an example, our model showed an 11\% improvement in MSE and a 9.7\% improvement in MAE at a predicted length of 24. When the predicted length was 336, MSE increased by 10.36\% and MAE increased by 6.0\%. It can be seen that with the increase of the predicted length, our model still has a good improvement. Since we added stacked GCN model in Informer, the training time will be about 2 minutes slower than that of Informer model. This is because in adaptive graph convolution networks, we need to perform additional matrix operations, as well as the iteration of the adjacency matrix. In the Autoformer model, we also achieved an improvement in accuracy with the addition of adaptive GCN, which is the highest in the current model.


\section{Conclusions}
In this paper, We proposed a new framework, which we named ADPGCN. Our model can adaptively capture the hidden dependencies between different dimensions and update the adjacency matrix with such dependencies, so that the GCN module can aggregate such dependencies and finally improve the accuracy. In addition, the coupling degree between the ADPGCN we added and the Informer model is very low. This model can also be added to other multi-dimensional time series prediction models. We've also demonstrated this by incorporating it into other models. In the future work,We'll explore whether this ADPGCN framework can be used to learn about other types of dependencies.

\appendices
%



\section*{Acknowledgment}

\ifCLASSOPTIONcaptionsoff
  \newpage
\fi


\bibliographystyle{IEEEtran}
\bibliography{mybibfile}
%

\end{document}